\documentclass[runningheads]{llncs}
\usepackage[T1]{fontenc}
\usepackage{graphicx,verbatim}
\usepackage{algorithm}
\usepackage{algpseudocode}
\usepackage{amsmath}
\usepackage{amssymb}
\usepackage{booktabs}
\usepackage{tabularx}
\usepackage{array}
\usepackage{svg}
\newcommand{\equalcontrib}{\textsuperscript{*}}
\newcommand{\corrauth}{\textsuperscript{\Letter}}
\usepackage{marvosym}
\begin{document}
\titlerunning{CW-B for Cardiac Phenotyping}
\title{CW-B: Class Weighted Boosting Framework for Imbalance Resilient Multi Class Cardiac Phenotyping}


\author{
Sijia Li\inst{1}\equalcontrib \and
Xiaoyu Tan\inst{2}\equalcontrib \and
Chen Zhan\inst{1} \and
Yuanji Ma\inst{3} \and
Haoyu Wang\inst{4} \and
Xihe Qiu\inst{1}\corrauth
}

\authorrunning{S. Li et al.}

\institute{
School of Electronic and Electrical Engineering, Shanghai University of Engineering Science, Shanghai 201620, China\\
\email{qiuxihe1993@gmail.com}
\and
Tencent YouTu Lab, Shanghai 200232, China
\and
Department of Cardiology, Zhongshan Hospital, Fudan University, Shanghai Institute of Cardiovascular Diseases, Shanghai 200032, China
\and
Department of Control Science and Engineering, College of Electronics and Information Engineering, Tongji University, Shanghai 200092, China
}
\maketitle
\begingroup
\renewcommand{\thefootnote}{}
\footnotetext{\textsuperscript{*}Equal contribution. \quad \Letter~Corresponding author.}
\endgroup

\begin{abstract}
Cardiac discharge phenotyping informs post-discharge treatment and follow-up, but real-world records are often incomplete and class-imbalanced, increasing the risk of missed high-risk phenotypes. We propose \textbf{CW-B}, a clinical risk-aligned class-weighted XGBoost pipeline for five-class cardiac discharge phenotyping under real-world class imbalance and missingness. CW-B combines fold-specific class-balanced instance weighting, missingness-indicator augmentation, and classwise error auditing to improve recognition of clinically prioritized phenotypes while preserving interpretable and auditable decision logic. In five-fold stratified cross-validation, CW-B achieves the best Accuracy, Macro-F1, Balanced Accuracy, and Prioritized F1 among tree-based, ensemble, and neural baselines. Overall, CW-B provides a practical and deployment-oriented approach for more reliable cardiac discharge phenotyping in real-world clinical settings.

\end{abstract}

\section{Introduction}

Automated phenotyping of hospital discharge diagnoses serves as a critical bridge between in-hospital care and post-discharge management, directly informing risk stratification, therapeutic planning, and longitudinal follow-up pathways~\cite{hripcsak2013next}. Compared with single-outcome prediction, discharge phenotyping more closely matches real clinical decision entry points and therefore requires models to produce stable and consistent assignments under substantial physiological heterogeneity, incomplete documentation, and noisy measurements \cite{denaxas2019phenotyping}. Moreover, the clinical consequences of errors are inherently asymmetric, as missed identification of acute high-risk phenotypes can divert subsequent management and incur disproportionately severe harm. Consequently, algorithms for discharge phenotyping should achieve strong overall discrimination while explicitly auditing and reducing high-risk missed detections, supported by interpretable and auditable evidence to facilitate clinical adoption.

Prior work in clinical phenotyping and computational phenomics spans multiple methodological paradigms \cite{banda2018advances,alzoubi2019review}. Rule-based and expert systems offer transparency but often depend on institution-specific coding practices and heuristic thresholds, limiting generalizability across populations and sites \cite{gehrmann2017comparison}. Conventional machine learning and ensemble models are effective on structured data \cite{japkowicz2002class}, yet under skewed class distributions they frequently optimize global metrics in ways that underemphasize clinically salient phenotypes \cite{lewis2023electronic,he2009learning}. More recently, end-to-end deep models and large language models have been explored for EHR representation learning and clinical reasoning; however, their failure modes remain difficult to characterize, and limitations in interpretability and auditability continue to impede deployment \cite{yang2023machine,de2021phe2vec}. Overall, there remains a need for a practical framework that explicitly models class bias and asymmetric risk while providing a controllable trade-off and reproducible evidence suitable for clinical auditing \cite{ding2024identify,doshi2017towards}.

To address this gap, we propose \textbf{CW-B}, a risk-aligned class-weighted boosting pipeline for cardiac discharge phenotyping.
Rather than treating imbalance handling as a generic preprocessing step, CW-B integrates fold-specific class-balanced instance weighting, explicit missingness modeling, and clinically prioritized error auditing within a unified five-class prediction task. In addition, CW-B employs robust preprocessing and explicit modeling of missingness to improve resilience to incomplete records and measurement variability commonly observed in real-world data. While retaining the computational efficiency and deployment practicality of tree-based models, CW-B preserves interpretability through feature-based split structures, supporting traceable decision rationales and systematic auditing \cite{caruana2015intelligible}.

Our contributions are threefold:
\begin{enumerate}
 \item We formulate cardiac discharge phenotyping as a risk-aligned five-class task with explicit emphasis on clinically actionable missed detections.
 \item We implement CW-B with fold-specific class-balanced instance weighting and leakage-free missingness-aware feature construction.
 \item We benchmark CW-B against tree-based, ensemble, and neural baselines using global metrics and prioritized classwise error auditing.
\end{enumerate}

\begin{figure}[t]
  \centering
  \includegraphics[width=\textwidth]{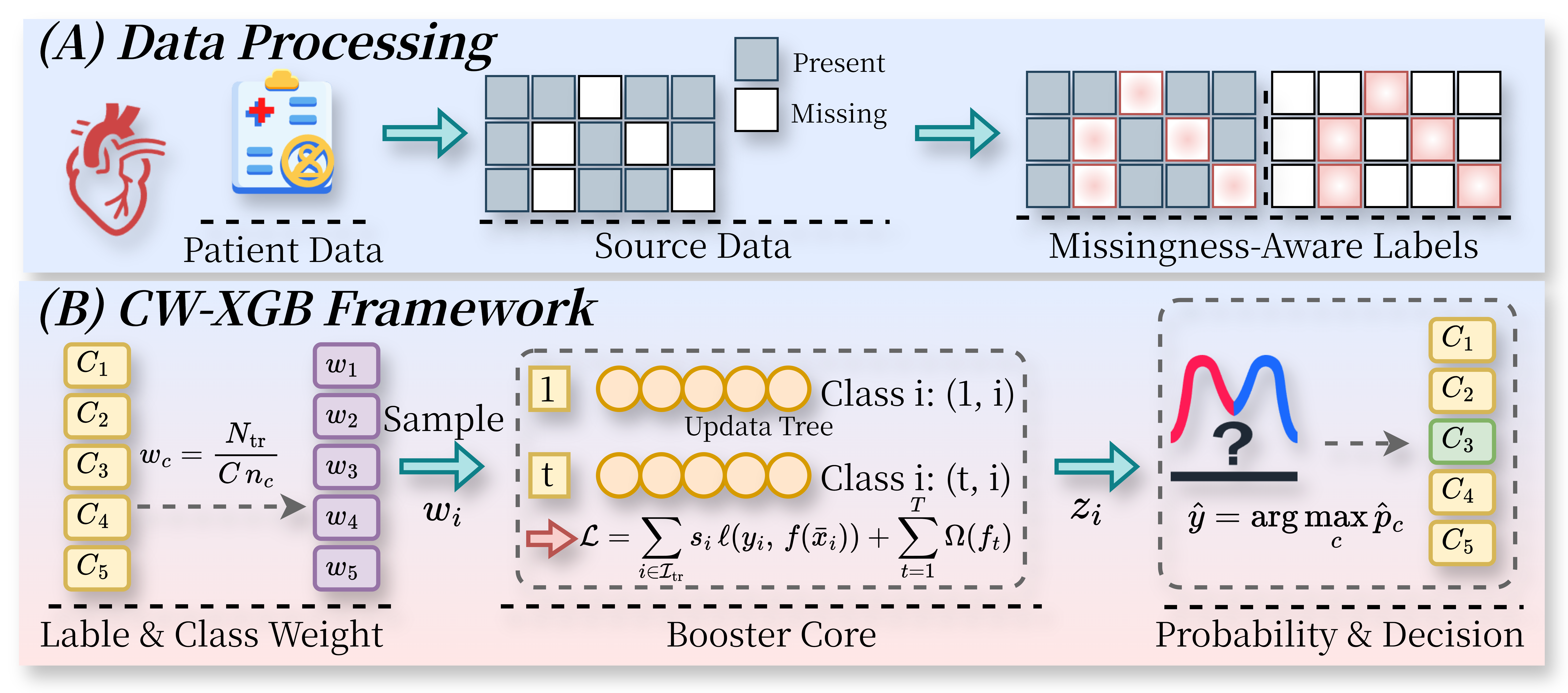}
  \caption{Overview of the proposed CW-B pipeline. (A) Data processing and construction of missingness-aware labels, where binary missingness indicators encode whether each clinical variable is originally observed or missing. (B) The CW-XGB classifier core, which performs class-weighted training and probability-based multiclass prediction.}
  \label{fig:cwxgb_overview}
\end{figure}

\section{Methods}

\subsection{Task Definition and Notation}
We study discharge phenotyping as a multi class classification problem over structured clinical records. An overview of the overall pipeline and the proposed CW-B framework is shown in Fig.~\ref{fig:cwxgb_overview}.
Let
\begin{equation}
\mathcal{D}=\{(x_i,y_i)\}_{i=1}^{N}, \qquad x_i\in\mathbb{R}^{d}, \qquad y_i\in\{0,\ldots,C-1\},
\end{equation}
where $x_i$ is the feature vector and $y_i$ is the discharge phenotype label. In our main setting, $C=5$ with the label mapping
\begin{equation}
0=\mathrm{stableCAD},\quad 1=\mathrm{ACS},\quad 2=\mathrm{oldMI},\quad 3=\mathrm{CAS},\quad 4=\mathrm{nonCAD}.
\end{equation}
Here, \textbf{stableCAD} (label 0) refers to stable coronary artery disease with chronic or stable ischemic presentation; \textbf{ACS} (label 1) denotes acute coronary syndrome characterized by time-sensitive acute ischemic events; \textbf{oldMI} (label 2) indicates a prior myocardial infarction documented as historical disease; \textbf{CAS} (label 3) denotes non-obstructive coronary artery disease and, in our cohort, corresponds to clinically suspected CAD patients from the younger, initially non-CAD subgroup who underwent coronary angiography demonstrating $<50\%$ luminal narrowing. Disease burden was further quantified using the Gensini score (GS), and patients were followed longitudinally to monitor incident CAD-related events \cite{reynolds2023ischemia,widmer2019functional}; and \textbf{nonCAD} (label 4) represents patients without a CAD-related discharge phenotype. In clinical practice, \textbf{0/1/3} (stableCAD, ACS, and CAS) are associated with established diagnostic and therapeutic pathways, making them clinically actionable categories for downstream management, which motivates our emphasis on these classes in deployment-oriented interpretation while retaining a unified five-class learning objective.

The model outputs a probability vector $p_i=f(x_i)\in[0,1]^C$ and a prediction $\hat{y}_i=\arg\max_c p_{ic}$. Before model evaluation, we define the clinically prioritized set
\begin{equation}
\mathcal{C}_p=\{0,1,3\},
\end{equation}
corresponding to stableCAD, ACS, and CAS. This set is pre-defined based on clinical actionability and missed-detection risk rather than class frequency or experimental results. Specifically, ACS is a time-sensitive acute ischemic phenotype, stableCAD informs secondary prevention, medication planning, and follow-up, and CAS in this cohort refers to clinically suspected CAD patients with less than 50\% luminal narrowing who still require risk assessment and symptom management. Importantly, $\mathcal{C}_p$ is used only for deployment-oriented evaluation and classwise error auditing; it does not change the training labels, model objective, or prediction rule.

\begin{algorithm}[t]
\caption{Class weighted gradient boosted trees for one stratified fold}
\label{alg:cwxgb}
\begin{algorithmic}[1]
\Require Indices $\mathcal{I}_{\mathrm{tr}},\mathcal{I}_{\mathrm{ev}}$, features $X\in\mathbb{R}^{N\times d}$, labels $y\in\{0,\ldots,C-1\}^N$, classes $C$, hyperparameters $\Theta$, missing sentinel $s_{\mathrm{miss}}$
\Ensure Fold model $f_k$, preprocessing parameters $\Pi_k$, predictions $(\hat{y},\hat{p})$
\State $r \leftarrow \mathbb{I}(X=s_{\mathrm{miss}})$
\State Set $X_{ij}\leftarrow \mathrm{NaN}$ wherever $r_{ij}=1$
\State Compute $(\mu,\sigma,m)$ on $X[\mathcal{I}_{\mathrm{tr}}]$ and set $\Pi_k=(\mu,\sigma,m)$
\State Standardize using $(\mu,\sigma)$ and impute missing values using $m$
\State Form $\bar{X}_{\mathrm{tr}}=[\tilde{X}_{\mathrm{tr}};\,r[\mathcal{I}_{\mathrm{tr}}]]$ and $\bar{X}_{\mathrm{ev}}=[\tilde{X}_{\mathrm{ev}};\,r[\mathcal{I}_{\mathrm{ev}}]]$
\State $n_c \leftarrow \sum_{i\in\mathcal{I}_{\mathrm{tr}}}\mathbb{I}(y_i=c),\ \forall c$
\State $w_c \leftarrow \frac{|\mathcal{I}_{\mathrm{tr}}|}{C\,n_c},\ \forall c$
\State $\tilde{w}_c \leftarrow \frac{w_c}{\frac{1}{C}\sum_j w_j},\ \forall c$
\State Assign $s \leftarrow \tilde{w}_{y[\mathcal{I}_{\mathrm{tr}}]}$
\State Train $f_k$ on $(\bar{X}_{\mathrm{tr}},y[\mathcal{I}_{\mathrm{tr}}])$ with instance weights $s$ and hyperparameters $\Theta$
\State Predict probabilities $\hat{p}$ on $\bar{X}_{\mathrm{ev}}$ and set $\hat{y}=\arg\max_c \hat{p}_{c}$
\State \Return $f_k,\Pi_k,(\hat{y},\hat{p})$
\end{algorithmic}
\end{algorithm}

\subsection{Preprocessing and Explicit Missingness Modeling}
We treat preprocessing as an integral component of the learning pipeline and enforce a leakage free protocol. Missing entries in the raw feature table are encoded by a sentinel value $s_{\mathrm{miss}}$ and are converted to missing values prior to transformation. All preprocessing statistics are estimated using the training split only and then applied to the corresponding evaluation split \cite{perez2022benchmarking,sterne2009multiple}.

Given a training index set $\mathcal{I}_{\mathrm{tr}}$, each feature dimension $j$ is standardized using the training mean $\mu_j$ and standard deviation $\sigma_j$,
\begin{equation}
x'_{ij}=\frac{x_{ij}-\mu_j}{\sigma_j}.
\end{equation}
To ensure numerical stability, invalid or very small $\sigma_j$ values are stabilized to a positive constant during fitting. Missing entries are imputed using the training median $m_j$,
\begin{equation}
\tilde{x}_{ij}=
\begin{cases}
m_j, & \text{if } x'_{ij} \text{ is missing},\\
x'_{ij}, & \text{otherwise}.
\end{cases}
\end{equation}
Because missingness itself can be informative, we additionally construct a binary indicator $r_{ij}\in\{0,1\}$ that represents whether feature $j$ is missing for sample $i$ \cite{che2018recurrent,gillies2020demonstrating}. In our cohort, 55 of 57 features contain missing values, with 38,571 missing entries among 248,178 total feature entries, corresponding to a 15.54\% missingness rate. This motivates preserving documentation patterns through explicit missingness indicators, allowing the model to distinguish an imputed numerical value from the clinical fact that the value was originally unrecorded. The final augmented representation concatenates imputed values and missingness indicators,
\begin{equation}
\bar{x}_i=\bigl[\tilde{x}_i;\,r_i\bigr]\in\mathbb{R}^{2d},
\end{equation}
allowing the model to jointly leverage observed values and recording patterns under incomplete documentation.

\subsection{Class Balanced, Instance Weighted Gradient Boosted Trees}
Our primary model is a class balanced, instance weighted gradient boosted decision tree classifier \cite{chen2016xgboost}. The objective is to mitigate label imbalance by controlling the contribution of each class during learning while preserving transparent, auditable tree structured decision rules \cite{haixiang2017learning}.

Let $n_c$ denote the number of training samples in class $c$ and $N_{\mathrm{tr}}$ denote the number of training samples. We define class balanced weights
\begin{equation}
w_c=\frac{N_{\mathrm{tr}}}{C\,n_c}, \qquad
\tilde{w}_c=\frac{w_c}{\frac{1}{C}\sum_{j=0}^{C-1}w_j}, \qquad
s_i=\tilde{w}_{y_i},
\end{equation}
where $s_i$ is the instance weight assigned to sample $i$. This construction reduces the dominance of frequent phenotypes and improves class balanced learning.

Let $\ell(\cdot)$ denote the multiclass negative log likelihood and let $\{f_t\}_{t=1}^{T}$ be the additive sequence of trees with regularizer $\Omega(\cdot)$. The weighted learning objective is
\begin{equation}
\mathcal{L}=\sum_{i\in\mathcal{I}_{\mathrm{tr}}}s_i\,\ell\!\left(y_i,\, f(\bar{x}_i)\right)+\sum_{t=1}^{T}\Omega(f_t).
\end{equation}
Optimization follows gradient boosting with a second order approximation, where instance weights modulate both split selection and leaf value estimation through weighted gradients and curvatures \cite{zadrozny2003cost}.

\section{Experiments}

\begin{table}[t]
\small
\centering
\caption{Hyperparameter configuration for the proposed method and baselines.}
\label{tab:hyperparams_en}
\setlength{\tabcolsep}{4pt}
\renewcommand{\arraystretch}{1.08}
\begin{tabularx}{\textwidth}{@{}p{3.0cm}X@{}}
\toprule
\textbf{Method} & \textbf{Key settings} \\
\midrule
\textbf{CW-B (Ours)} &
600 boosting iterations, depth 5, learning rate 0.07, subsampling 0.95, $\ell_2$ 1.0 \\
XGB &
1200 boosting iterations, depth 7, learning rate 0.03, subsampling 0.95, $\ell_2$ 1.0 \\
CB &
1200 boosting iterations, depth 8, learning rate 0.03 \\
STK &
CatBoost and XGBoost base learners, logistic regression meta learner, out of fold training \\
Neural baselines &
MLP(256) for DQN and BC, MLP(256,128) with early stopping; Adam learning rate $10^{-3}$ \\
\bottomrule
\end{tabularx}
\end{table}

\subsection{Evaluation Protocol}
We evaluate the proposed method on a five-class discharge phenotyping task with $N=4354$ samples and $d=57$ structured features. The class counts for stableCAD, ACS, oldMI, CAS, and nonCAD are 1675, 483, 266, 461, and 1469, respectively. Stratified five-fold cross-validation is used with a fixed random seed \cite{kohavi1995study}. Preprocessing statistics are estimated on the training split only and then applied to the corresponding evaluation split, ensuring that evaluation data do not influence feature construction. All component ablations are conducted under the same five-fold stratified cross-validation protocol, random seed, and main model configuration unless otherwise specified.

We report overall classification accuracy (Accuracy), macro averaged F1 score (Macro-F1), and balanced accuracy (Balanced Accuracy), where Balanced Accuracy is defined as the average of per class recall \cite{brodersen2010balanced,opitz2022bias}. To emphasize deployment relevant failure modes, we additionally report averaged recall, averaged F1 score, and averaged miss rate over the clinically prioritized set $\mathcal{C}_p=\{0,1,3\}$, corresponding to stableCAD, ACS, and CAS.

\begin{table}[t]
\small
\centering
\caption{Mean CV performance. Prioritized metrics average stableCAD, ACS, and CAS; lower is better for miss rate.}
\label{tab:main_compare_en}
\setlength{\tabcolsep}{3.2pt}
\renewcommand{\arraystretch}{1.10}
\begin{tabularx}{\textwidth}{@{}>{\raggedright\arraybackslash}p{2.2cm}*{7}{>{\centering\arraybackslash}X}@{}}
\toprule
\textbf{Meth} &
\textbf{Acc} &
\textbf{M-F1} &
\shortstack{\textbf{Bal}\\\textbf{Acc}} &
\shortstack{\textbf{Pri}\\\textbf{Rec}} &
\shortstack{\textbf{Pri}\\\textbf{F1}} &
\shortstack{\textbf{Pri}\\\textbf{Miss}} &
\shortstack{\textbf{M-F1}\\\textbf{Std}} \\
\midrule
\textbf{CW-B(ours)} & \textbf{0.81} & \textbf{0.72} & \textbf{0.73} & \underline{0.67} & \textbf{0.69} & \underline{0.33} & 0.0103 \\
XGB         & \underline{0.80} & 0.69 & 0.66 & 0.65 & \underline{0.67} & 0.35 & 0.0108 \\
CB          & 0.78 & \underline{0.70} & 0.70 & 0.66 & 0.66 & 0.34 & 0.0136 \\
STK         & 0.75 & 0.69 & \underline{0.71} & \textbf{0.69} & 0.66 & \textbf{0.31} & 0.0069 \\
DQN         & 0.76 & 0.68 & 0.68 & 0.66 & 0.64 & 0.34 & 0.0070 \\
BC          & 0.77 & 0.68 & 0.67 & 0.62 & 0.64 & 0.38 & 0.0109 \\
MLP         & 0.76 & 0.65 & 0.63 & 0.58 & 0.60 & 0.42 & 0.0180 \\
\bottomrule
\end{tabularx}
\end{table}

\subsection{Baselines and Hyperparameter Configuration}
We benchmark CW-B against tree based, ensemble, and neural baselines under the same preprocessing constraints and the same evaluation protocol. XGB is an XGBoost baseline with a higher capacity configuration, using more boosting iterations and deeper trees to assess whether gains are attributable to capacity rather than the proposed weighting strategy. CB is a CatBoost baseline that evaluates an alternative gradient boosting implementation under class balancing \cite{prokhorenkova2018catboost}. STK is a stacking ensemble that combines CatBoost and XGBoost base learners and fits a logistic regression meta learner on out of fold predictions, providing a controlled test of whether model fusion can improve clinically prioritized errors beyond single learners \cite{naimi2018stacked}. DQN and BC are multilayer perceptron baselines trained with cost sensitive objectives based on focal loss and class balancing, serving as neural references that emphasize minority sensitivity \cite{pomerleau1988alvinn,foster2024behavior}. MLP is a standard multilayer perceptron trained with early stopping, serving as a non cost sensitive neural comparator \cite{lin2017focal,mnih2015human}.

These baselines are selected to isolate and validate the key components of the proposed approach. Tree based baselines (XGB, CB) test whether class balanced instance weighting and the explicit missingness representation yield consistent gains under strong tabular inductive bias \cite{grinsztajn2022tree}. The higher capacity XGBoost configuration tests whether the proposed benefits persist beyond capacity scaling. The stacking ensemble tests whether complementary decision boundaries can further reduce missed detections in the prioritized phenotypes. Neural baselines test whether cost sensitive training alone is sufficient to match boosted trees on this structured, moderately sized cohort, thereby supporting the choice of a transparent tree based model for deployment. Table~\ref{tab:hyperparams_en} summarizes the core hyperparameter configuration and imbalance handling strategy.

\begin{figure}[t]
\centering
\includegraphics[width=\linewidth]{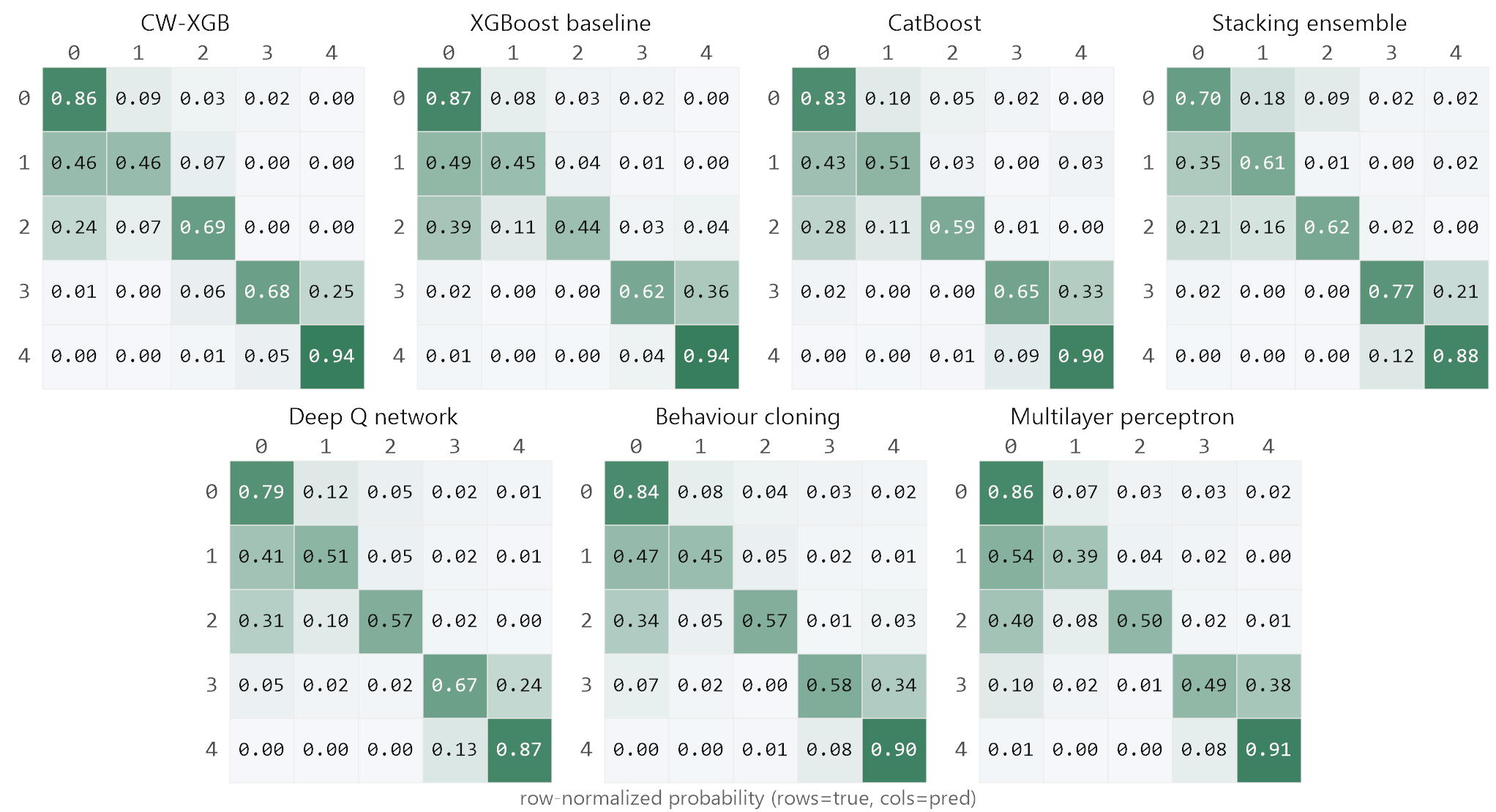}
\caption{Concatenated confusion matrices. Each subplot corresponds to one method. Class order: stableCAD, ACS, oldMI, CAS, nonCAD.}
\label{fig:cm_en}
\end{figure}

\subsection{Metrics and Main Results}

Let $M\in\mathbb{N}^{C\times C}$ denote the confusion matrix, where rows represent true labels and columns represent predicted labels. We report Accuracy, Macro-F1, and Balanced Accuracy using their standard definitions, with Balanced Accuracy computed as the mean class-wise recall. For class $c$, let $R_c$ denote recall and $\mathrm{F1}_c$ denote the class-wise F1 score. Specifically,
\begin{equation}
R_c=\frac{M_{cc}}{\sum_{b=0}^{C-1}M_{cb}},
\qquad
\mathrm{MissRate}_c=1-R_c.
\end{equation}
For the clinically prioritized set $\mathcal{C}_p=\{0,1,3\}$, corresponding to stableCAD, ACS, and CAS, we further report
\begin{equation}
\begin{aligned}
\mathrm{Prioritized\ Recall}
&=\frac{1}{|\mathcal{C}_p|}\sum_{c\in\mathcal{C}_p}R_c, \\
\mathrm{Prioritized\ F1}
&=\frac{1}{|\mathcal{C}_p|}\sum_{c\in\mathcal{C}_p}\mathrm{F1}_c, \\
\mathrm{Prioritized\ MissRate}
&=1-\mathrm{Prioritized\ Recall}.
\end{aligned}
\end{equation}

Figure~\ref{fig:cm_en} visualizes the concatenated confusion matrices, enabling inspection of systematic confusions involving stableCAD, ACS, and CAS. Table~\ref{tab:main_compare_en} reports mean performance for all methods, sorted by Macro-F1, then by the prioritized F1, and then by Accuracy.

\subsection{Discussion}
CW-B achieves the best Accuracy, Macro-F1, Balanced Accuracy, and Prioritized F1 among the compared methods, while remaining competitive on prioritized recall and miss rate. As shown in Table~\ref{tab:main_compare_en}, CW-B improves over the higher-capacity XGBoost baseline, suggesting that the gain is not simply attributable to model capacity. Additional component analyses under the same five-fold stratified cross-validation protocol further support the contribution of the main design choices. Removing class-balanced instance weighting reduces Macro-F1, Balanced Accuracy, and Prioritized F1 to 0.6968, 0.6621, and 0.6631, respectively. Removing the missingness indicator yields 0.6804, 0.6666, and 0.6444, while relying only on XGBoost native missing-value handling yields 0.7082, 0.6909, and 0.6780. Alternative imputation with indicators also underperforms the full pipeline; for example, KNN imputation yields 0.7008, 0.6791, and 0.6703. These results indicate that both class-balanced learning and explicit missingness modeling contribute to macro-level and prioritized-class performance. Fig.~\ref{fig:cm_en} provides complementary qualitative evidence by visualizing classwise error patterns. The stacking ensemble achieves the lowest prioritized miss rate, but this sensitivity-oriented operating point comes with a clear trade-off in overall accuracy and macro-level balance. In contrast, CW-B preserves strong global performance while delivering the best prioritized F1 for the physician-emphasized phenotypes.

\section{Conclusion}
We propose CW-B, a class weighted gradient boosted tree framework with explicit missingness modeling for imbalance resilient multi class discharge phenotyping. In stratified cross validation, CW-B achieves the strongest macro level performance with competitive overall accuracy and balanced accuracy, and it yields more clinically coherent confusion patterns with fewer clinically undesirable errors. Clinically, these improvements are meaningful because discharge phenotypes directly guide secondary prevention, medication selection, follow up intensity, and referral pathways, and more reliable phenotype assignments can reduce downstream management variability and support earlier, more consistent risk aligned care after hospitalization. In addition, the transparent tree based structure facilitates post hoc auditing and error review, which is essential for building clinical trust and enabling safe integration into real world discharge workflows.

\section*{Disclosure of Interests}
The authors have no competing interests to declare that are relevant to the content of this article.

\bibliographystyle{splncs04}
\bibliography{refs}

\end{document}